\PassOptionsToPackage{table}{xcolor}

\documentclass[10pt, a4paper, logo]{googledeepmind}

\usepackage{amsmath,amsfonts,bm}
\usepackage{natbib}
\usepackage{fancyhdr}

\def\eqref#1{equation~\ref{#1}}

\def\1{\bm{1}}

\DeclareMathAlphabet{\mathsfit}{\encodingdefault}{\sfdefault}{m}{sl}
\SetMathAlphabet{\mathsfit}{bold}{\encodingdefault}{\sfdefault}{bx}{n}

\usepackage{subcaption}
\usepackage{siunitx}
\usepackage{natbib}
\usepackage{array}
\usepackage{multirow}
\usepackage{xspace}
\usepackage{algorithm}
\usepackage{algpseudocode}
\usepackage{verbatim}
\usepackage[normalem]{ulem}
\usepackage{wrapfig}
\usepackage{etoolbox}

\tcbuselibrary{breakable}

\titleformat{\paragraph}[runin]
{\normalsize\bfseries}
{}
{0em}
{#1}
[]

\newcommand{\method}{NapMem\xspace}

\title{From Passive Retrieval to Active Memory Navigation: Learning to Use Memory as a Structured Action Space}

\author[1,2]{Yue Xu\textsuperscript{*}}
\author[3]{Yutao Sun\textsuperscript{*}}
\author[4]{Yihao Liu\textsuperscript{*}}
\author[1]{Mengyu Zhou\textsuperscript{$\dag$}}
\author[5]{Jiayi Qiao\textsuperscript{*}}
\author[1]{Lu Ma}
\author[1]{Kai Tang}
\author[2]{Wenjie Wang\textsuperscript{$\dag$}}
\author[1]{Xiaoxi Jiang}
\author[1]{Guanjun Jiang}
\affil[1]{Qwen Large Model Application Team, Alibaba}
\affil[2]{ShanghaiTech University}
\affil[3]{Zhejiang University}
\affil[4]{Peking University}
\affil[5]{National University of Singapore}
\footnotetext{\textsuperscript{*}Work done during an internship at Alibaba. \textsuperscript{$\dag$}Corresponding author.}

\correspondingauthor{wangwj1@shanghaitech.edu.cn, zhoumengyu.zmy@alibaba-inc.com}

\begin{abstract}
Long-term user memory is essential for personalized conversational agents, yet many memory systems still expose memory through passive retrieval interfaces, making the model a consumer of pre-selected evidence. We introduce \method, a framework for learning to use long-term user memory as a structured action space rather than passively retrieved context. \method organizes user history into a linked multi-granularity memory pyramid, where raw conversations, typed memory records, topic tracks, and user profiles are connected through provenance relations, and exposes these levels through memory tools. The agent is trained to select memory according to the query and intermediate evidence, allowing it to inspect different memory granularities before answering. Experiments on PersonaMem-v2, LongMemEval, and LoCoMo show that a \method agent trained with memory-tool reinforcement learning is competitive across diverse memory-intensive tasks, while evaluations on non-memory tasks suggest that the learned policy largely preserves general reasoning and tool-use abilities. Additional analyses examine storage, inference cost, tool-use behavior, and ablations over navigation, memory granularity, and RL training. Our results suggest that long-term user memory benefits from coupling structured storage with a learned policy for using memory at the appropriate granularity.
\end{abstract}

\begin{document}
\maketitle

\section{Introduction}

\begin{wrapfigure}{r}{0.45\textwidth}
    \centering
    \vspace{-3.5em}
    \includegraphics[width=0.42\textwidth]{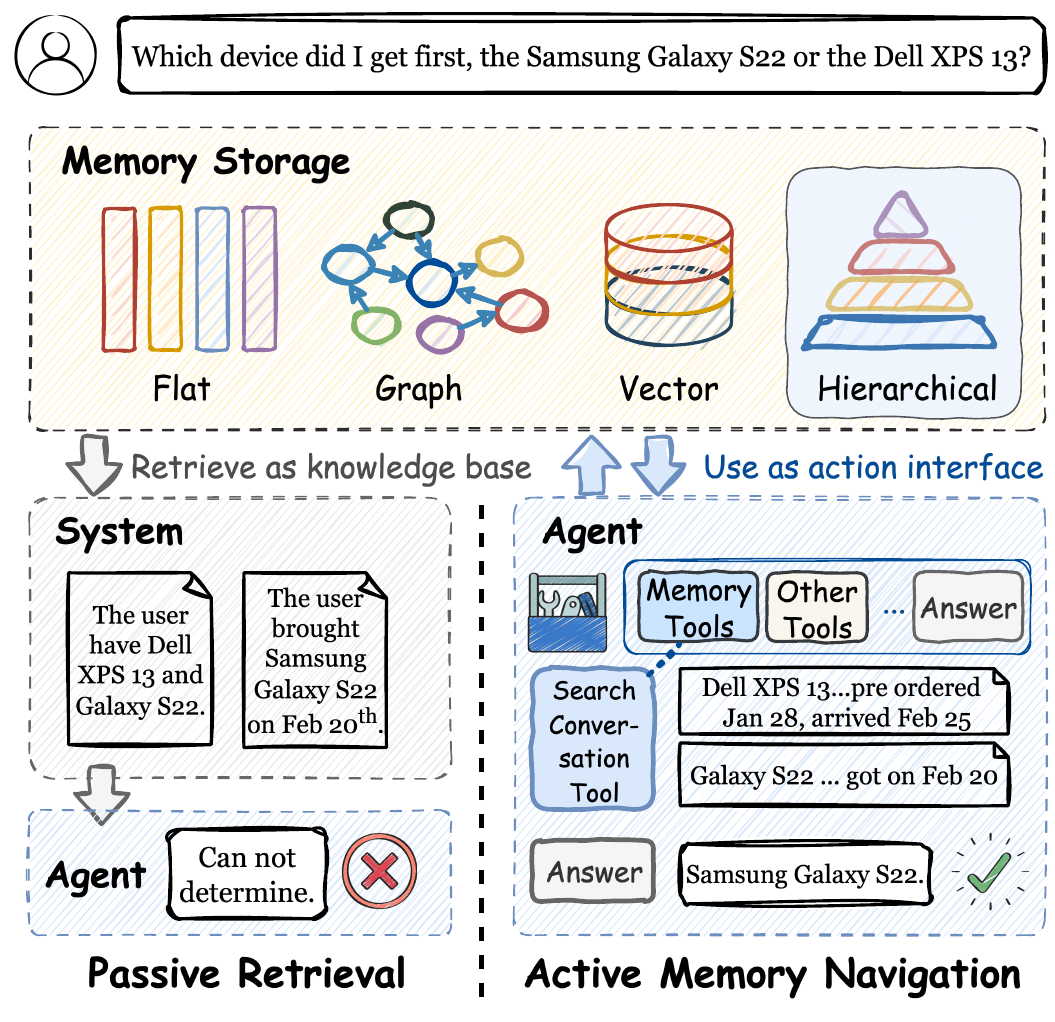}
    \vspace{-1em}
    \caption{Motivating example for active memory navigation. Passive retrieval leaves the agent with partial evidence, while \method exposes long-term user memory as an action interface, enabling the agent to search for sufficient evidence before answering.}
    \label{fig:teaser}
    \vspace{-2em}
\end{wrapfigure}

As Large Language Model (LLM) agents are deployed in personal assistance, education, workplace productivity, and other long-running interactive settings, they are expected to maintain continuity and personalization across sessions, requiring agents to build a persistent understanding of the user. Long-term user memory is therefore becoming an essential component for personalized conversational agents \citep{hu2025memory,packer2024memgptllmsoperatingsystems,xu2026toward}. However, building an effective user memory system is challenging since user information is multifaceted and future queries impose heterogeneous memory requirements \citep{zhong2023memorybankenhancinglargelanguage}. Different queries may require different types of user memory, making it difficult for a fixed memory representation or access strategy to generalize across future interactions.

Existing efforts improve user memory from two directions: memory construction and memory retrieval. On the construction side, recent systems build richer memory infrastructures by designing finer-grained memory categories, introducing vector, graph, hybrid, or hierarchical storage structures, and using agentic methods to manage memory entries \citep{chhikara2025mem0buildingproductionreadyai,rasmussen2025zeptemporalknowledgegraph,li2025memosoperatingmemoryaugmentedgeneration,kang2025memoryosaiagent}. On the retrieval side, another line of work makes memory access more adaptive by introducing reflective retrieval, personalizing retrieval queries, or optimizing retrieval reranking \citep{tan2025prospect,jiang2025deepretrieval,zhang2026personalize}. However, these approaches largely treat memory access as a system-level retrieval function or a predesigned pipeline, leaving the agent with limited control over how memory is utilized.

We argue that \textbf{\textit{long-term user memory use should move from system-level passive retrieval to agent-native memory navigation}}. As illustrated in Figure~\ref{fig:teaser}, passive retrieval reduces different memory storage structures to pre-selected context and delivers it to the agent, limiting the agent's ability to inspect additional evidence when the retrieved context is incomplete. 
In contrast, active memory navigation formulates long-term user memory as a structured action space and equips the agent with memory tools, allowing it to decide whether memory is needed, which level of abstraction to consult, and whether the acquired evidence is sufficient. We instantiate this view with \textbf{\method} (\underline{\textbf{Na}}vigate over \underline{\textbf{P}}yramid \underline{\textbf{Mem}}ory), a framework that trains agents to navigate memory as a structured action space.

\begin{figure*}[t]
    \centering
    \includegraphics[width=\textwidth]{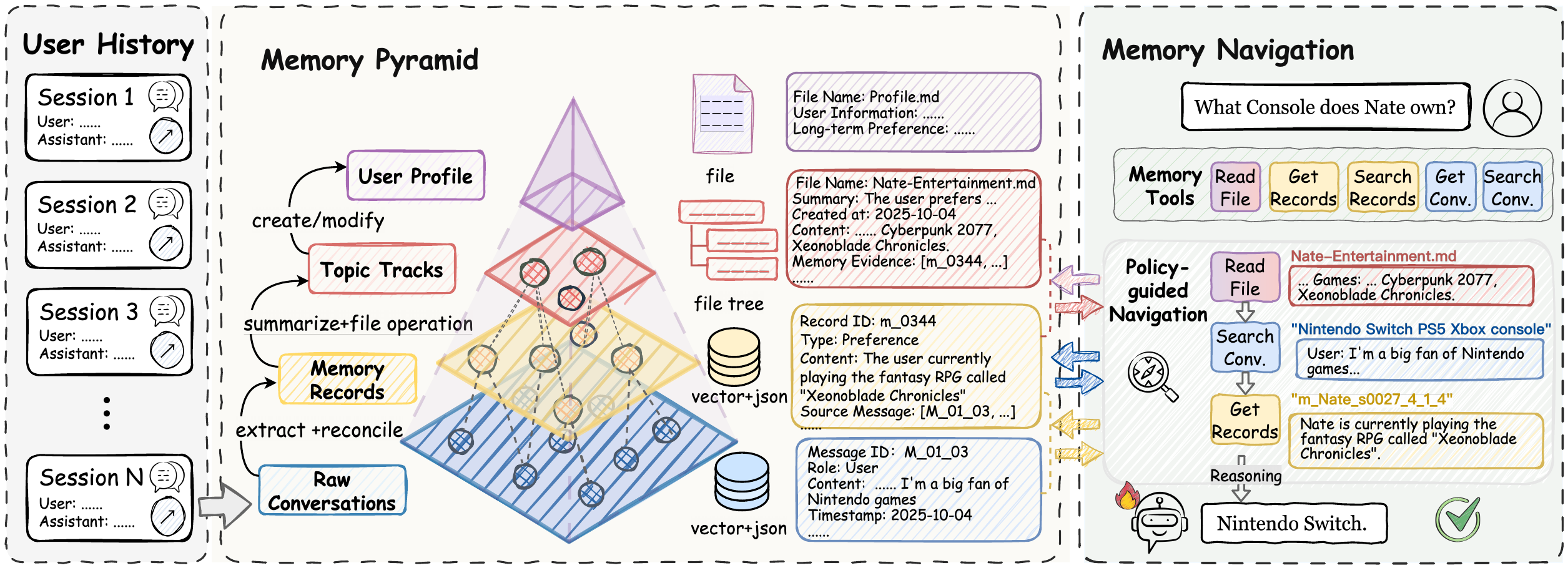}
    \caption{Overview of \method. The agent actively navigates a multi-granularity long-term memory pyramid through memory tools, selecting appropriate abstraction levels and refining actions based on intermediate evidence.}
    \label{fig:main}
\end{figure*}

Specifically, \method first organizes user interaction histories into a multi-granularity memory pyramid. As shown in Figure~\ref{fig:main}, the pyramid spans raw conversations for evidence-level verification, typed memory records for compact facts and preferences, topic tracks for cross-session aggregation, and user profiles for stable user-level information. These levels are linked through provenance relations, making the memory bank a navigable interface over different levels of user understanding.
To enable navigation over this interface, \method exposes memory access through tools tailored to different memory granularities. The agent is then trained with reinforcement learning to use these tools according to the query and the evidence gathered so far. In this way, memory use becomes an explicit and optimizable part of the agent's decision process: the agent can start from a broad summary, drill down to specific evidence, or stop once enough information has been collected.


We evaluate \method on six benchmarks covering both long-term user-memory tasks and non-memory tasks. Across three memory-intensive benchmarks, \method achieves competitive performance on diverse user-memory requirements, including personalization, long-range recall, and cross-session reasoning. Across three non-memory benchmarks, \method largely preserves performance on selected reasoning and tool-use tasks, suggesting that memory-use training does not necessarily compromise broader agent capabilities. Ablations indicate that the gains are associated with the coupling between the memory pyramid and the learned navigation policy, while tool-call analyses reveal query-dependent memory-use behaviors, offering insights toward more agent-native use of long-term memory. Our contributions can be summarized as follows:
\begin{itemize}[leftmargin=*, itemsep=0.2em, topsep=0.2em, parsep=0pt, partopsep=0pt]
    \item We formulate long-term user-memory use as active memory navigation, reframing memory from passively retrieved context into a structured action space that agents can learn to use.
    \item We introduce \method, a framework that builds a multi-granularity memory pyramid and exposes raw conversations, memory records, topic tracks, and user profiles through tools for controllable memory access.
    \item We train an agent to navigate this structured memory action space with reinforcement learning over memory-tool trajectories. 
    \item We evaluate \method across memory-intensive and non-memory benchmarks, analyzing task performance, generalization, efficiency, and tool-use behavior.
\end{itemize}

\section{Related Work}

\paragraph{Memory construction and organization.}
A large body of work studies how long-term memories should be constructed, organized, and maintained for LLM agents. Early user-memory systems such as MemoryBank and MemGPT introduce persistent user history and virtual context management for long-running interactions \citep{zhong2023memorybankenhancinglargelanguage,packer2024memgptllmsoperatingsystems}. Recent systems further explore different memory architectures: Mem0 constructs compact memory records for scalable retrieval, Zep organizes memory as a temporal knowledge graph, MemOS treats memory as a system-level resource for memory-augmented generation, and MemoryOS proposes a hierarchical short-/mid-/long-term personal memory architecture \citep{chhikara2025mem0buildingproductionreadyai,rasmussen2025zeptemporalknowledgegraph,li2025memosoperatingmemoryaugmentedgeneration,kang2025memoryosaiagent}. Other recent work improves memory organization from complementary perspectives. A-MEM dynamically links and evolves memories in an agentic manner, and LightMem and SimpleMem emphasize lightweight compression and efficient consolidation \citep{xu2026mem,fang2026lightmem,liu2026simplemem}.

\paragraph{Adaptive retrieval for user memory.}
Another line of work improves the retrieval stage. Reflective Memory Management (RMM) proposes retrospective reflection, which refines retrieval through online reinforcement learning based on cited evidence \citep{tan2025prospect}. More broadly, retrieval-oriented training methods such as Search-R1 and DeepRetrieval optimize query generation, enabling models to produce more effective search queries during reasoning or retrieval \citep{jin2025searchr1,jiang2025deepretrieval}. \citet{zhang2026personalize} further shows that user-specific signals can improve retrieval by expanding queries according to user expression style and corpus structure before retrieval. These methods make retrieval more adaptive through query generation, while the downstream agent still follows a fixed memory-use interface.

\paragraph{RL-based and agentic memory systems.}
Learning-based memory systems have recently explored how to replace hand-crafted memory routines with trainable policies. One line of work focuses on memory construction, training agents to decide what information should be stored, how memories should be structured, or when they should be updated using downstream task feedback \citep{wang2025memalpha,kang2026memreader}. Another line treats memory management as an operation-selection problem, where agents learn or orchestrate actions such as writing, updating, deleting, summarizing, or retrieving memories across long-term and short-term stores \citep{yan2026memoryr1,yu2026agenticmemory,zhang2026memskill}. Recent routing-based systems further externalize memory planning by assigning queries to different memory paths or modules \citep{chen2026memflow}. These studies show that memory behavior can be learned or orchestrated beyond fixed heuristics. \method instead studies query-time use of a structured user-memory pyramid, training a single conversational agent to select memory levels and refine access based on intermediate evidence.

\section{\method}

This section presents \method, a framework that trains agents to navigate memory as a structured action space. We first provide an overview of the framework, then describe the construction of the multi-layer memory pyramid, the five memory tools tailored to different memory granularities, and the training objective for optimizing the memory-navigation policy.

\subsection{Overview}
As shown in Figure \ref{fig:main}, given a user's interaction history, \method constructs a multi-granularity memory pyramid that organizes user information from raw conversational evidence to high-level user profiles.
At inference time, the agent interacts with the memory pyramid through memory tools and sequentially selects which memory level to access. We optimize this memory-use behavior with Group Relative Policy Optimization (GRPO) \citep{shao2024deepseekmath}, using rewards that promote accurate answers, valid format, and appropriate memory use under a tool-call budget.

\subsection{Multi-Granularity Memory Pyramid}
\method constructs a separate memory pyramid for each user by processing the user's interaction history in chronological order. The construction is incremental and bottom-up: newly observed sessions are first appended to the raw-conversation layer and then used to extract and reconcile memory records, which may further update topic tracks and the user profile. In this way, lower-layer changes are propagated upward when new evidence extends, supersedes, or contradicts existing memories. Each level provides a different degree of abstraction over the same user history, and adjacent levels are connected through provenance links that allow the agent to move between detailed conversational evidence and summarized user understanding during inference. In implementation, the two lower layers are stored as structured JSONL records with unique identifiers and auxiliary vector and keyword indices, enabling exact lookup and hybrid retrieval. The two upper layers are stored as Markdown files, making them directly readable through memory tools. See Algorithm~\ref{alg:memory_construction} in Appendix~\ref{app:memory_construction} for details.

\paragraph{Raw conversations.}
The base layer stores the original user-agent interactions as message-level entries, each containing the raw content, speaker role, timestamp, and a unique identifier for exact lookup. This append-only layer preserves the highest-fidelity evidence, allowing the agent to recover fine-grained details when needed.

\paragraph{Memory records.}
The second layer converts raw conversational evidence into compact memory records. This layer serves as the basic semantic unit of user memory, abstracting away conversational redundancy while preserving retrievable and verifiable user information. We define four record types to cover different aspects of user information: \textit{fact}, \textit{event}, \textit{instruction}, and \textit{preference}. The record layer is updated through a two-stage incremental procedure. First, after a batch of complete user-agent turns has accumulated, the agent receives the newly appended conversations together with recent records as background and produces new records supported by the current dialogue. Each record follows a fixed schema, including a unique record identifier, type, textual content, creation or update time, source message identifiers, and other type-specific metadata. Second, newly extracted records are reconciled with the existing related record bank through hybrid retrieval. This step removes redundant records, updates outdated or conflicting information, and merges related records when appropriate. The reconciled records are registered with persistent record identifiers and indexed for later access, while retaining update history and source message identifiers. These identifiers provide a bridge from compact user-memory records back to raw conversational evidence.

\paragraph{Topic tracks.}
The third layer provides medium-range abstraction by organizing related records into evolving user-centric topic tracks. While memory records capture atomic information, many dimensions of user understanding develop across sessions and require longitudinal tracking over multiple pieces of evidence. A topic track maintains the accumulated context around a recurring user topic, and is updated through an agent-driven writing process. Given newly created or revised records, the agent determines whether to update an existing track, merge related tracks, or start a new track. Each track is stored as a file with metadata, a short summary, and a narrative description of the topic. Its content is grounded with explicit links to supporting memory record identifiers, enabling topic tracks to capture cross-session evolution and recurring user contexts while preserving a downward path to the record layer.

\paragraph{User profile.}
The top layer maintains a global user profile that summarizes stable user attributes, long-term preferences and interaction patterns. The profile is initialized after the first session and updated when sufficient new records or topic-track changes have accumulated. During each update, the profile writer revises the existing profile file under a length budget, preserving a compact set of long-term user insights. This layer provides a concise global view of the user for personalization and high-level contextualization.

\subsection{Tool-Based Memory Navigation}
\method exposes memory pyramid levels through a set of memory tools, turning memory use into a sequential navigation process over the pyramid. The tool interface makes memory access explicit, requiring the agent to decide which memory operation to invoke based on the user query and the information gathered so far.

Formally, given a user query $q$ and a memory pyramid $\mathcal{M}$, the agent sequentially interacts with memory before producing the final answer. At each step $t$, the agent either invokes a memory tool or terminates navigation and answers:
\[
a_t \sim \pi_\theta(\cdot \mid q, \mathcal{M}, a_{<t}, o_{<t}),
\]
where $a_t$ denotes the next action and $o_t$ denotes the observation returned when a memory tool is invoked. The resulting memory-use trajectory is
$$
\tau = (a_1, o_1, \ldots, a_k, o_k, y),
$$
where $y$ is the final answer.

\method is equipped with five tools tailored to different memory layers: \texttt{get\_conversations}, \texttt{search\_ conversations}, 
\texttt{get\_records}, \texttt{search\_records}, and \texttt{read\_files}. Search tools retrieve candidate raw conversation snippets or memory records through hybrid search. Get tools access exact messages or records through their persistent identifiers. The file-reading tool allows the agent to inspect topic tracks and the user profile, which provide higher-level abstractions of user history. These tools support both top-down and bottom-up navigation: the agent may begin with global user context, refine its search through topic- or record-level memories, or descend to raw conversations for evidence-level verification.

\subsection{Learning Memory Navigation with GRPO}
We optimize the memory-navigation policy with Group Relative Policy Optimization (GRPO) \citep{shao2024deepseekmath}. During training, each sample consists of a memory-intensive user query paired with the corresponding memory pyramid, and the agent is required to answer the query under a maximum tool-call budget. We assign only a terminal reward to each trajectory, so that tool-use decisions and final-answer generation are optimized with respect to the same task-level outcome.

\paragraph{Reward rubric.}
We use a rule-based reward over three binary criteria: format validity $F$, answer correctness $C$, and memory-tool usage $U$. Here, $F$ indicates whether the final answer satisfies the required answer format, $C$ indicates whether the answer is correct, and $U$ indicates whether the trajectory invokes at least one memory tool. Since RL training uses memory-intensive queries, memory-tool usage is treated as a desired behavior during training. Trajectories that exceed the tool-call budget or fail to produce a valid final answer are treated as invalid outputs. The reward is defined as
\[
r(\tau)=
\left\{
\begin{array}{ll}
-1, & F=0,\\
\phantom{-}1, & F=1,\ C=1,\ U=1,\\
\phantom{-}0, & F=1,\ C=1,\ U=0,\\
-0.5, & F=1,\ C=0,\ U=1,\\
-1, & F=1,\ C=0,\ U=0.
\end{array}
\right.
\]
For multiple-choice tasks, correctness is computed by exact accuracy; for open-ended tasks, it is determined by an LLM judge following the corresponding benchmark protocol. This reward assigns the highest score to trajectories that answer correctly through memory use, while penalizing malformed outputs and incorrect answers.

\paragraph{Group-relative optimization.}
For each query, let $\mathcal{G}$ denote the group of sampled trajectories and let $r_i=r(\tau_i)$ be the terminal reward of trajectory $\tau_i$. We compute the group-relative advantage as
\[
A_i = r_i - \frac{1}{|\mathcal{G}|}\sum_{j \in \mathcal{G}} r_j .
\]
The policy is optimized with the clipped GRPO objective:
\[
\mathcal{J}(\theta)
=
\mathbb{E}\left[
\frac{1}{|\mathcal{G}|}\sum_{i=1}^{|\mathcal{G}|}
\frac{1}{T_i}\sum_{t=1}^{T_i}
\ell_{i,t}(\theta)
-
\beta D_{\mathrm{KL}}(\pi_\theta \| \pi_{\mathrm{ref}})
\right],
\]
where
\[
\ell_{i,t}(\theta)
=
\min\left(
\rho_{i,t} A_i,\,
\mathrm{clip}(\rho_{i,t}, 1-\epsilon, 1+\epsilon)A_i
\right).
\]
Here, $\rho_{i,t}$ is the token-level importance ratio, $T_i$ is the number of generated tokens in trajectory $\tau_i$. The trajectory-level advantage is applied uniformly to all output tokens, including tokens that specify memory-tool calls. As a result, the terminal reward jointly optimizes final-answer quality and the preceding memory-navigation behavior.

\section{Experiments}

\subsection{Experiment Setup}

\paragraph{Datasets.}
We evaluate \method on six benchmarks covering both long-term user-memory tasks and non-memory tasks. For \textbf{memory-intensive tasks}, we use PersonaMem-v2 \citep{jiang2025personamemv2personalizedintelligencelearning}, LongMemEval \citep{wu2025longmemevalbenchmarkingchatassistants}, and LoCoMo \citep{maharana-etal-2024-evaluating}. These benchmarks cover two complementary aspects of long-term user memory. LongMemEval and LoCoMo focus on user-specific factual memory, including long-range recall, temporal changes, and cross-session reasoning. PersonaMem-v2 focuses on implicit user preferences and persona understanding, evaluating whether the agent can personalize its responses based on user-specific behavioral signals.
For \textbf{non-memory tasks}, we evaluate whether memory-use training preserves broader agentic capabilities and avoids inducing unnecessary memory calls. We consider general reasoning with GPQA-Diamond (GPQA-D) \citep{rein2023gpqa}, text-based function calling with the Berkeley Function Calling Leaderboard (BFCL-v3) \citep{patil2025bfcl}, and visual tool-use question answering with V*Bench \citep{Wu_2024_CVPR}. Detailed dataset descriptions are provided in Appendix \ref{app:dataset}.
  
\paragraph{Metrics.} For open-ended memory QA tasks such as LoCoMo and LongMemEval, we report F1 and LLM-judge accuracy (L-J). For multiple-choice tasks such as PersonaMem-v2 and GPQA-Diamond, we report accuracy. For BFCL-v3 and V*Bench, we report task success or accuracy following their standard evaluation protocols.

\paragraph{Baselines and models.}
We compare \method against five representative memory baselines: Mem0 \citep{chhikara2025mem0buildingproductionreadyai}, Zep \citep{rasmussen2025zeptemporalknowledgegraph}, MemOS \citep{li2025memosoperatingmemoryaugmentedgeneration}, MemoryOS \citep{kang2025memoryosaiagent}, and AgeMem \citep{yu2026agenticmemory}. These baselines cover both memory architectures and agentic memory policies: Mem0 represents flat, vector-based memory; Zep represents temporal graph memory; MemOS represents an agent-oriented memory operating system; MemoryOS represents a hierarchical memory architecture; and AgeMem represents an RL-trained agentic memory baseline for unified memory management and use. All baselines use Qwen3.5-9B \citep{qwen35blog} as the base LLM, and our default \method further trains this model with the proposed memory-navigation framework. Unless otherwise specified, \method refers to this RL-trained 9B model. In the main comparison, we additionally report two untrained scaling variants, \method-122B and \method-397B, which use Qwen3.5-122B and Qwen3.5-397B as base models without training.

\paragraph{Implementation details.}
We train \method with memory-tool reinforcement learning on memory-intensive queries sampled from LoCoMo and PersonaMem-v2. For each training dataset, we use a 60/20/20 user-level split for training, validation, and testing to avoid user-level leakage. LongMemEval is used as an out-of-distribution memory benchmark, while the held-out splits of LoCoMo and PersonaMem-v2 evaluate in-domain generalization. During inference, \method may invoke memory tools for at most four steps per query; trajectories exceeding this budget are terminated and counted as incorrect. Each retrieval operation in \method returns up to five memory items. AgeMem follows the same inference-time tool-step budget. For other non-agentic baselines, each retrieval operation is allowed to return up to twenty memory items, while system-specific settings otherwise follow the original implementations. We use Qwen3-Embedding-0.6B \citep{yang2025qwen3} as the shared encoder for all embedding-based retrieval components. For open-ended memory QA, we use \texttt{Claude-Opus-4.6} as the judge model and conduct human checks to validate the reliability of its judgments. All reported results are averaged over three random seeds. Additional implementation details, hyperparameters, and prompt templates are provided in Appendices~\ref{app:implementation} and~\ref{app:prompt}.

\subsection{Main Comparison}
\paragraph{Results on long-term user-memory tasks.}
Table~\ref{tab:main} reports results on the three memory-intensive benchmarks. \method-9B w/ RL obtains the highest average score of 62.74, compared with 59.85 for \method-397B. On LongMemEval and PersonaMem-v2, our method achieves the best scores among the evaluated systems. On LoCoMo, it achieves the best F1 score and is close to the strongest L-J score. These results show that the trained 9B model performs competitively across different user-memory settings, including factual recall, cross-session reasoning, and implicit preference understanding. Long-term user-memory performance benefits not only from structured memory storage but also from a trained policy for selecting and using the appropriate memory actively.

\begin{table*}[!ht]
  \centering
  \small
  \renewcommand{\arraystretch}{1.1}
  \resizebox{\textwidth}{!}{
  \begin{tabular}{lcccccc}
  \toprule
  \multirow{2}{*}{\textbf{Method}} & \multicolumn{2}{c}{\textbf{LoCoMo}} & \multicolumn{2}{c}{\textbf{LongMemEval}} & \textbf{PersonaMem-v2} & \multirow{2}{*}{\textbf{Avg.}} \\
  \cmidrule(lr){2-3} \cmidrule(lr){4-5} \cmidrule(lr){6-6}
  & F1 & L-J & F1 & L-J & Acc. & \\
  \midrule
  Mem0 & 41.19 $\pm$ 0.36 & \textbf{60.86 $\pm$ 0.38} & \underline{53.86} $\pm$ 0.54 & 78.00 $\pm$ 0.82 & 38.89 $\pm$ 0.34 & 59.25 $\pm$ 0.32 \\
  Zep & 36.09 $\pm$ 0.38 & 52.21 $\pm$ 0.31 & 52.61 $\pm$ 0.20 & 73.33 $\pm$ 0.94 & 36.33 $\pm$ 0.81 & 53.96 $\pm$ 0.43 \\
  MemOS & 35.22 $\pm$ 0.38 & 55.82 $\pm$ 0.65 & 39.16 $\pm$ 0.40 & 54.33 $\pm$ 0.47 & 37.58 $\pm$ 0.44 & 49.24 $\pm$ 0.30 \\
  MemoryOS & 31.40 $\pm$ 0.53 & 44.28 $\pm$ 1.22 & 22.68 $\pm$ 0.30 & 23.67 $\pm$ 0.58 & 35.82 $\pm$ 1.87 & 34.59 $\pm$ 0.05 \\
  AgeMem & 38.00 $\pm$ 0.12 & 45.02 $\pm$ 0.12 & 37.83 $\pm$ 1.38 & 51.33 $\pm$ 0.94 & 24.19 $\pm$ 0.67 & 40.18 $\pm$ 0.37 \\
  \midrule
  \method-122B & 35.39 $\pm$ 0.26 & 51.69 $\pm$ 0.84 & 48.86 $\pm$ 2.58 & 70.00 $\pm$ 1.73 & 41.38 $\pm$ 0.77 & 54.36 $\pm$ 0.74 \\
  \method-397B & 38.31 $\pm$ 0.13 & 54.42 $\pm$ 0.81 & 53.85 $\pm$ 1.25 & \underline{78.33} $\pm$ 2.08 & \underline{46.10} $\pm$ 1.15 & \underline{59.85} $\pm$ 1.14 \\
  \rowcolor[HTML]{EBF5EC} \method-9B w/ RL & \textbf{41.28 $\pm$ 0.59} & \underline{59.92} $\pm$ 0.78 & \textbf{57.41 $\pm$ 1.12}& \textbf{80.33 $\pm$ 0.47} & \textbf{47.97 $\pm$ 1.01} & \textbf{62.74 $\pm$ 0.45}
  \\
  \bottomrule
  \end{tabular}
  }
  \caption{Memory-intensive task performance on LoCoMo, LongMemEval, and PersonaMem-v2. L-J denotes the LLM-judge accuracy. Avg. is the mean of LoCoMo L-J, LongMemEval L-J, and PersonaMem-v2 accuracy.}
  \label{tab:main}
\end{table*}

\paragraph{\method preserves non-memory agent capabilities.}
We evaluate non-memory benchmarks to test whether memory-tool training induces unnecessary memory access. For each query, we randomly assign a user memory pyramid and enable all memory tools, while the task itself does not require user history. Table~\ref{tab:generalization} shows that \method-9B w/ RL remains competitive with the base model on non-memory tasks and improves several metrics on GPQA-D and V*Bench. Compared with the no-RL variant, RL substantially reduces unnecessary memory calls on GPQA-D, from 34.51\% to 6.90\%, and keeps memory calls at zero on BFCL-v3 and V*Bench. This behavior indicates that RL helps calibrate memory-tool use rather than simply increasing memory access, while largely preserving non-memory reasoning and tool-use performance.

\begin{table}[!ht]
\centering
\small
\begin{tabular}{lcccc}
\toprule
\multirow{2}{*}{\textbf{Benchmark}} & \multirow{2}{*}{\textbf{Metric}} &\multirow{2}{*}{\textbf{Base}} & \multicolumn{2}{c}{\textbf{\method-9B}} \\
\cmidrule(lr){4-5}
& & & w/o RL & w/ RL\\
\midrule
GPQA-D & Acc. & 53.03 & 55.74& \textbf{57.58} \\
 & Memory call & -- & 34.51 & 6.90 \\
\midrule
BFCL-v3 & Task acc. & 95.00 & 93.50 & 95.00 \\
        & Tool acc. & 97.50 & 97.50 & 97.50 \\
        & Exec acc. & 93.50 & 94.00& \textbf{94.50} \\
        & Memory call & -- & \textbf{0.00} & \textbf{0.00} \\
\midrule
V*Bench & Answer Acc. & 84.82 &89.53 & \textbf{91.10} \\
   & Tool acc. & 97.91 & 98.95 & \textbf{99.48}\\
   & Exec acc. & 97.91 & 98.95 & \textbf{99.48}\\
   & Memory call & -- & \textbf{0.00} & \textbf{0.00}\\
\bottomrule
\end{tabular}
\caption{Generalization to non-memory tasks. We compare the base model and \method on reasoning and tool-use benchmarks, with memory-tool call rates reported to measure unnecessary memory access.}
\label{tab:generalization}
\end{table}

\subsection{Efficiency}

\paragraph{Storage efficiency.}
Storage efficiency is important for long-term user memory because user histories grow continuously over time. Table~\ref{tab:storage} shows that \method uses less storage than most memory-system baselines despite exposing multiple memory granularities. Although AgeMem has a smaller footprint due to its lightweight flat-memory design, its task performance is substantially lower in Table~\ref{tab:main}. This suggests that storage size alone is not sufficient. By separating raw evidence, compact records, and higher-level abstractions, \method provides a lightweight structure for scalable long-term memory while preserving the information needed for flexible memory navigation.

\begin{table*}[h]
\centering
\small
\setlength{\tabcolsep}{5pt}
\begin{tabular}{lcccccc}
\toprule
\textbf{Dataset} & \textbf{\method} & \textbf{Mem0} & \textbf{MemOS} & \textbf{Zep} & \textbf{MemoryOS} & \textbf{AgeMem} \\
\midrule
\textbf{PersonaMem-v2} 
& 0.69 
& 1.40{\scriptsize~(2.03$\times$)} 
& 3.61{\scriptsize~(5.23$\times$)} 
& 3.50{\scriptsize~(5.07$\times$)} 
& 1.98{\scriptsize~(2.87$\times$)} 
& 0.34{\scriptsize~(0.49$\times$)} \\

\textbf{LongMemEval} 
& 4.01 
& 8.90{\scriptsize~(2.22$\times$)} 
& 18.85{\scriptsize~(4.70$\times$)} 
& 10.00{\scriptsize~(2.49$\times$)} 
& 4.30{\scriptsize~(1.07$\times$)} 
& 2.56{\scriptsize~(0.64$\times$)} \\

\textbf{LoCoMo} 
& 0.12 
& 0.14{\scriptsize~(1.17$\times$)} 
& 0.64{\scriptsize~(5.33$\times$)} 
& 0.60{\scriptsize~(5.00$\times$)} 
& 0.40{\scriptsize~(3.33$\times$)} 
& 0.09{\scriptsize~(0.75$\times$)} \\
\midrule
\textbf{Total} 
& 4.83 
& 10.44{\scriptsize~(2.16$\times$)} 
& 23.10{\scriptsize~(4.78$\times$)} 
& 14.10{\scriptsize~(2.92$\times$)} 
& 6.68{\scriptsize~(1.38$\times$)} 
& 2.99{\scriptsize~(0.62$\times$)} \\
\bottomrule
\end{tabular}
\caption{Memory-bank storage size in GiB. Values in parentheses denote storage relative to \method.}
\label{tab:storage}
\end{table*}

\begin{figure}[!ht]
    \centering
    \vspace{-.5em}
    \includegraphics[width=0.45\textwidth]{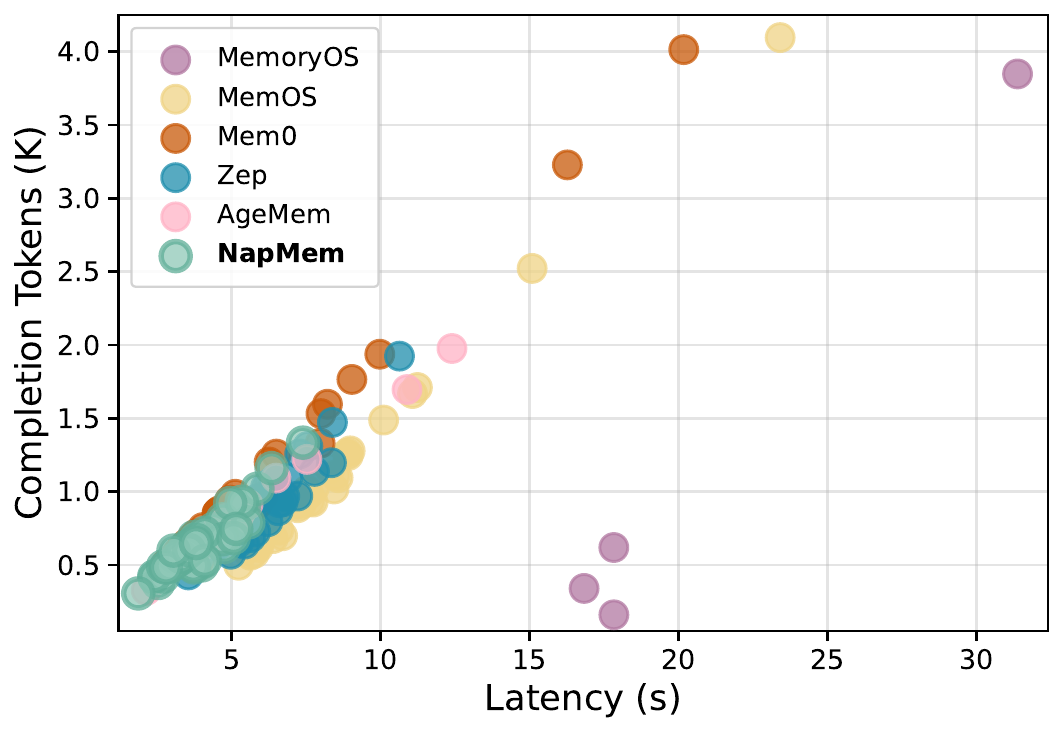}
    \vspace{-.5em}
    \caption{Latency and completion-token statistics on successful runs from 100 PersonaMem-v2 samples.}
    \vspace{-1em}
    \label{fig:efficiency}
\end{figure}

\paragraph{Inference efficiency.}
Inference efficiency is important for long-term conversational agents, where memory access must improve personalization without increasing interaction latency or response verbosity. Figure~\ref{fig:efficiency} reports inference efficiency on 100 samples, with statistics computed over successful runs for each method. The scatter plot compares average latency and average completion tokens and shows an approximately linear relationship between the two, suggesting that generation length is a major contributor to inference cost. Compared with baselines that passively reason over retrieved memory, \method performs targeted memory navigation and stops once sufficient evidence is gathered, leading to shorter completions and lower latency despite using memory tools.

\subsection{Ablation Study}

We perform ablations to examine the contributions of active navigation, memory granularity, and RL training. Table~\ref{tab:ablation} reports the results of four variants on the three memory-intensive benchmarks. As shown, \textbf{\textit{w/o navigation}} replaces tool-based memory access with passive retrieval while keeping the underlying memory sources available. \textbf{\textit{records-only tools}} preserves the tool interface but restricts the agent to compact memory records. \textbf{\textit{w/o upper levels}} allows access to raw conversations and memory records but removes topic tracks and the user profile. \textbf{\textit{w/o RL}} keeps the same memory pyramid and memory tools as Full \method, but uses the base model without reinforcement learning.
The results suggest that all three components contribute to performance. \textit{w/o navigation} falls behind Full \method, showing the value of query-conditioned tool use over passive retrieval. The lower scores of \textit{records-only tools} and \textit{w/o upper levels} suggest that higher-level abstractions complement low-level evidence access. Finally, the gap between \textit{w/o RL} and Full \method indicates that the agent benefits from learning how to use the memory interface, rather than relying on prompting alone.

\begin{table}[!ht]
\centering
\small
\resizebox{.48\textwidth}{!}{
\begin{tabular}{lcccc}
\toprule
\textbf{Variant} & \textbf{LCM.} & \textbf{LME.} & \textbf{PMem.} & \textbf{Avg.} \\
\midrule
\textbf{Full \method} & \textbf{59.92} & \textbf{80.33} & \textbf{47.97} & \textbf{62.74} \\
\quad \textit{w/o RL} & 52.09 & 72.33 & 20.75 & 48.39 \\
\quad \textit{w/o navigation} & 48.37 & 68.00 & 45.88 & 54.08 \\
\quad \textit{records-only tools} & 39.77 & 60.00 & 35.02 & 44.93 \\
\quad \textit{w/o upper levels} & 49.13 & 75.00 & 38.20 & 54.11 \\
\bottomrule
\end{tabular}}
\caption{Ablation study on memory access design. We report LLM-judge accuracy for LoCoMo and LongMemEval, and accuracy for PersonaMem-v2.}
\label{tab:ablation}
\end{table}

\vspace{-.5em}

\subsection{Tool-Use Behavior Analysis}

\begin{figure}[!ht]
    \centering
    \includegraphics[width=0.6\textwidth]{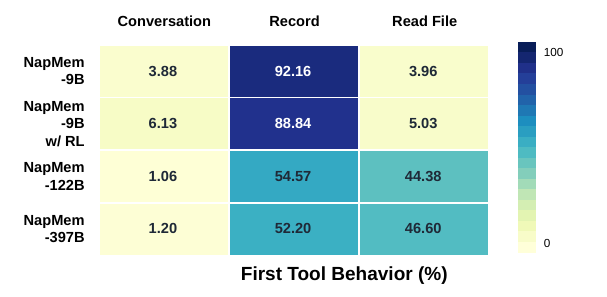}
    \vspace{-1em}
    \caption{First-step memory-tool behavior across \method variants.}
    \label{fig:tool_behavior}
    \vspace{-1em}
\end{figure}

\paragraph{Navigation strategies vary with model scale.}
We first analyze the first memory action to understand how different models initiate memory navigation. As shown in Figure~\ref{fig:tool_behavior}, larger \method variants more often start with file reading, while the 9B variants mostly start from record-level tools. This suggests that larger models tend to begin with broader topic-track or profile information, while smaller models prefer localized evidence such as records or conversations. One possible explanation is that larger models can better absorb long-form memory files, whereas smaller models benefit from starting with shorter, more targeted evidence.

\paragraph{RL improves navigation efficiency.}
Table~\ref{tab:tool_behavior} shows that RL training makes memory navigation more selective. Multi-level navigation rate denotes the fraction of samples that use more than one memory level, while evidence-hit ratio denotes the number of returned tool results that match supporting evidence per 100 tool uses. The trained model uses fewer tool calls while achieving higher overall accuracy, indicating that it learns to stop once sufficient evidence has been gathered. Meanwhile, the comparable multi-level navigation rate shows that this efficiency gain does not come from collapsing to a single memory level, and the higher evidence-hit ratio indicates more precise retrieval. 

\begin{table}[!ht]
\centering
\small
\resizebox{.48\textwidth}{!}{
\begin{tabular}{lrrr}
\toprule
Model & Tool Calls & Multi-level & Evidence Hit \\
\midrule
\method-9B w/ RL &\textbf{2.15} & 79.63\% & \textbf{34.92\%} \\
\method-9B & 3.97 & 80.39\% & 20.66\% \\
\method-122B & 2.50 & 82.34\% & 26.20\% \\
\method-397B & 2.69 & \textbf{88.22\%} & 26.42\% \\
\bottomrule
\end{tabular}}
\caption{Tool-use behavior across \method variants, including average tool calls, multi-level navigation rate, and evidence-hit ratio.}
\label{tab:tool_behavior}
\end{table}
\vspace{-1em}

\section{Conclusion}

We presented \method, a framework that reframes long-term user memory as active navigation over a multi-granularity memory pyramid. \method organizes user history into raw conversations, memory records, topic tracks, and user profiles, and trains agents to select the appropriate memory level through tool use. Experiments across memory-intensive and non-memory settings show that \method performs competitively on user-memory tasks while largely preserving general reasoning and tool-use abilities. Ablations and tool-use analyses further indicate that structured memory access and policy training help agents use memory more selectively. We hope \method encourages future work toward memory agents that treat memory use as an adaptive, learnable part of interaction.
\section*{Limitations}

\method focuses on making long-term user-memory use explicit and learnable. By organizing user history into a memory pyramid and exposing it through tools, the framework provides a controllable interface for studying how agents select memory granularity during inference.
This work has several limitations. First, the evaluation is conducted on existing long-term memory benchmarks, which may not capture all forms of real-world personalization and open-ended user interaction. Second, the current experiments do not fully address privacy and forgetting requirements that arise in user-memory systems. Third, broader scaling studies could further reveal how model size affects preferred memory-access strategies.

\bibliography{custom}
\bibliographystyle{conference}

\appendix
\clearpage
\section{Dataset Split Details}\label{app:dataset}

\paragraph{LoCoMo \citep{maharana-etal-2024-evaluating}.}
LoCoMo contains long multi-session conversations between two speakers, with questions that test long-term conversational memory. We convert each two-speaker conversation into speaker-specific user-agent memory banks by treating one speaker as the user and the other as the assistant, resulting in 20 users. We split users into train/validation/test with a 6/2/2 ratio. For each target speaker, we add a synthetic setup turn at the beginning of the memory bank, as shown below. If the first original utterance is already from the assistant/counterpart, only the user setup message is added. During evaluation, questions are asked from the corresponding speaker's perspective, and we remove questions whose evidence is missing or inconsistent with the provided answer in the original dataset. The final test set contains 1,315 questions.

\begin{tcolorbox}[
colback=orange!5,
colframe=orange!60,
title=LoCoMo speaker-specific setup turn,
fonttitle=\bfseries,
boxrule=0.5pt,
arc=2pt,
left=4pt,
right=4pt,
top=4pt,
bottom=4pt
]
\small
\textbf{User:} 
I am \{target\_speaker\}. You are \{counterpart\_speaker\}. Talk to me naturally like a friend. 
\medskip

\textbf{Assistant:} Got it.
\end{tcolorbox}

\paragraph{LongMemEval.} \citep{wu2025longmemevalbenchmarkingchatassistants}
LongMemEval evaluates long-term interactive memory across question types such as single-session recall, multi-session reasoning, temporal reasoning, knowledge updates, and preference-related questions. We use LongMemEval only as a held-out test benchmark to evaluate out-of-distribution generalization, with 100 test questions in total. 

\paragraph{PersonaMem-v2.} \citep{jiang2025personamemv2personalizedintelligencelearning}
PersonaMem-v2 evaluates personalized response selection from implicit user preferences and persona signals across long user-agent interactions. The benchmark contains 200 users, which we split by user into train/validation/test with a 6/2/2 ratio. The test set contains 911 questions. During evaluation, we randomly shuffle the answer options for each question to reduce positional bias.

\paragraph{GPQA-Diamond \citep{rein2023gpqa}.}
We use the GPQA-Diamond (GPQA-D) dataset to evaluate memory-independent scientific reasoning. It is a graduate-level, Google-proof multiple-choice benchmark containing difficult biology, physics, and chemistry questions written by domain experts.

\paragraph{V*Bench.} \citep{Wu_2024_CVPR}
We evaluate visual tool use on the official V* test set with 191 examples, including direct-attribute and relative-position questions. Each example contains an image, a multiple-choice visual question, and a ground-truth answer option. The model may call tools at most once and get the zoomed image as required for response.

\paragraph{BFCL-v3.} \citep{patil2025bfcl}
We evaluate text-based tool use on a 200-example executable subset of BFCL-v3, consisting of \texttt{exec\_simple}, \texttt{exec\_multiple}, and \texttt{exec\_parallel} examples. Each example provides a user request, executable function schemas, and ground-truth tool calls. The evaluator executes the model-generated tool calls with deterministic local backends and returns the results to the model. We report tool-call accuracy, execution accuracy, task accuracy, and unnecessary memory tool-call count.

\section{Memory Construction Details}
\label{app:memory_construction}

\method builds an independent memory pyramid for each user by replaying normalized sessions in chronological order. The builder appends raw messages, extracts and reconciles memory records, periodically updates topic tracks, and refreshes the user profile. Construction is incremental: lower-layer changes are propagated upward only when update triggers are met.

\begin{algorithm}[t]
\small
\caption{Incremental construction of a user memory pyramid.}
\label{alg:memory_construction}
\begin{algorithmic}[1]
\Require Chronologically sorted sessions $\mathcal{S}=\{s_1,\ldots,s_N\}$
\State Initialize raw store $\mathcal{C}$, record pool $\mathcal{R}$, pending buffer $\mathcal{B}$, topic tracks $\mathcal{T}$, profile $P$
\State Initialize topic and profile checkpoints $c_T,c_P \leftarrow 0$
\For{$s_i \in \mathcal{S}$}
\If{$i>1$ and \Call{TopicTrigger}{$\mathcal{B},c_T,s_{i-1},s_i$}}
    \State $\mathcal{T},c_T \leftarrow$ \Call{UpdateTopicTracks}{$\mathcal{T},\mathcal{R},\mathcal{B}$}
\EndIf
\If{$i>1$ and \Call{ProfileTrigger}{$\mathcal{T},P,c_P$}}
    \State $P,c_P \leftarrow$ \Call{UpdateProfile}{$P,\mathcal{T}$}
\EndIf

\State Append non-empty messages in $s_i$ to $\mathcal{C}$
\For{each completed turn batch $b$ in $s_i$}
    \State $\hat{\mathcal{R}} \leftarrow$ \Call{ExtractRecords}{$b$, recent context, recent records}
    \State $\tilde{\mathcal{R}} \leftarrow$ \Call{ReconcileRecords}{$\hat{\mathcal{R}},\mathcal{R}$}
    \State Add $\tilde{\mathcal{R}}$ to $\mathcal{R}$ and $\mathcal{B}$
    \If{\Call{TopicTrigger}{$\mathcal{B},c_T$}}
        \State $\mathcal{T},c_T \leftarrow$ \Call{UpdateTopicTracks}{$\mathcal{T},\mathcal{R},\mathcal{B}$}
    \EndIf
    \If{\Call{ProfileTrigger}{$\mathcal{T},P,c_P$}}
        \State $P,c_P \leftarrow$ \Call{UpdateProfile}{$P,\mathcal{T}$}
    \EndIf
\EndFor
\EndFor
\State Flush remaining records, update topic tracks and profile if triggered
\State Finalize active records, retrieval indices, topic index, and build metadata
\end{algorithmic}
\end{algorithm}

\paragraph{Raw conversations.}
For each non-empty message, \method stores the speaker role, content, timestamp, session identifier, message identifier, user identifier, and available metadata. Raw messages are also indexed for retrieval using embeddings. This layer is append-only and serves as the evidence source for exact lookup and verification.

\paragraph{Memory records.}
Records are extracted from recently completed user-agent turns. The flush schedule expands from $1$ to $2$ to $4$ turns and then uses a steady-state size of $5$ turns, with a final flush at session end. Each record contains a persistent identifier, content, type, source message identifiers, timestamps, session information, and update metadata. We use four record types: \textit{fact}, \textit{event}, \textit{instruction}, and \textit{preference}. New records are deduplicated against existing records using vector retrieval. The reconciliation model chooses one of four actions: \texttt{store}, \texttt{skip}, \texttt{update}, or \texttt{merge}. For \texttt{update} and \texttt{merge}, source message identifiers and timestamps are unioned to preserve provenance. The final records are indexed for later search.

\paragraph{Topic tracks.}
Topic tracks consolidate related memory records into medium-range user-centric narratives. Each track contains metadata, a concise summary, a narrative description, and explicit evidence links to supporting record identifiers. Topic updates are triggered before a new session when unprocessed records exist, at date boundaries, when more than $20$ new records have accumulated since the last topic checkpoint, at replay end, or during final retry recovery. The topic writer can inspect existing tracks, update or create tracks, perform localized edits, and finish the update round. The system maintains at most $20$ topic tracks. Updates are committed only after format validation; invalid writes are rolled back.

\paragraph{User profile.}
The profile summarizes stable user attributes, long-term preferences, interaction patterns, and high-level contextual information. It is initialized once topic tracks become available and is refreshed when requested by the topic writer, after successful topic-track updates, or when more than $50$ new records have accumulated since the last profile checkpoint. Each refresh revises the existing profile under a length budget. If validation fails, the previous profile is restored.

\section{Implementation Details}
\label{app:implementation}

We train PyraNav with multi-turn GRPO using the verl framework. At each step, the agent may either answer directly or invoke memory tools to inspect different levels of the memory bank. The memory bank is indexed
with Qwen3-Embedding-0.6B embeddings, and search tools use hybrid reciprocal-rank fusion over keyword and vector retrieval with $k=60$.

\subsection{RL Hyperparameters}

\begin{table}[!ht]
\centering
\small
\renewcommand{\arraystretch}{1.1}
\begin{tabular}{lc}
\toprule
\textbf{Hyperparameter} & \textbf{Value} \\
\midrule
RL algorithm & GRPO \\
Base model & Qwen3.5-9B \\
Train batch size & 8 \\
Rollout group size & 4 \\
Learning rate & $1\times 10^{-6}$ \\
Max assistant turns & 5 \\
KL loss coefficient & 0.001 \\
Entropy coefficient & 0 \\
Precision & bfloat16 \\
Total epochs & 5 \\
Hardware & NVIDIA H20 \\
\bottomrule
\end{tabular}
\caption{RL hyperparameters for \method-9B training.}
\label{tab:rl_hyperparameters}
\end{table}

\subsection{Tool Operation Protocol}
Table~\ref{tab:tool_operations} summarizes the memory tools exposed to the \method agent. The tools are aligned with different levels of the memory pyramid. Search tools support coarse retrieval over memory records and raw conversations, exact-fetch tools retrieve specific records or messages using identifiers returned by previous tool calls, and \texttt{read\_file} provides access to higher-level topic tracks and the user profile. 

\begin{table*}[!ht]
\centering
\small
\renewcommand{\arraystretch}{1.1}
\resizebox{\textwidth}{!}{
\begin{tabular}{lll}
\toprule
\textbf{Tool} & \textbf{Input} & \textbf{Functionality} \\
\midrule
\texttt{search\_records} & free-text query & Search structured memory records with hybrid RRF retrieval and return top-5 records. \\
\texttt{search\_conversation} & free-text query & Search raw conversation messages with hybrid RRF retrieval and return top-5 snippets. \\
\texttt{get\_records} & record id list & Fetch exact structured memory records by \texttt{record\_id}. \\
\texttt{get\_conversation} & message id list & Fetch exact raw conversation messages by \texttt{message\_id}. \\
\texttt{read\_file} & file name & Read a high-level memory-bank file, such as a topic-track file or \texttt{profile.md}. \\
\bottomrule
\end{tabular}}
\caption{Memory tools exposed to the agent. Search tools provide coarse retrieval, while exact-fetch and file-reading tools allow the model to navigate between memory granularities.}
\label{tab:tool_operations}
\end{table*}

\subsection{LLM Inference Hyperparameters}
For evaluation, we use the Qwen3-Coder tool-call format for parsing tool invocations. All the generations use temperature $1.0$, top-$p$ $0.9$, and a maximum completion length of $4096$ tokens. 

\subsection{Judge Validation}

For open-ended memory QA, we use \texttt{Claude-Opus-4.6} as the automatic judge. To validate the reliability of the judge, we manually annotate a random subset of 100 judged examples with three human annotators using the same prompt as LLM, and compare the LLM judgment against the human-majority label.

The LLM judge aligns closely with human-majority labels, achieving 99.0\% accuracy, 100.0\% precision, 98.0\% recall, 99.0\% F1, and Cohen's $\kappa=0.980$. The only disagreement is a false negative case where the gold answer indicates that the conversation did not mention baking egg tarts, while the model answered ``0 times''; the human majority considered this equivalent, but the LLM judge marked it incorrect. Human annotators also show high agreement: pairwise Cohen's $\kappa$ ranges from 0.920 to 0.960, and Fleiss' $\kappa$ across three annotators is 0.933. These results indicate that the LLM judge aligns with human-majority judgments at near-human reliability in our validation sample.
\begin{table}[t]
\centering
\small
\begin{tabular}{lc}
\toprule
\textbf{Metric} & \textbf{LLM vs. Human Majority} \\
\midrule
N & 100 \\
Accuracy & 0.990 \\
Precision & 1.000 \\
Recall & 0.980 \\
F1 & 0.990 \\
Cohen's $\kappa$ & 0.980 \\
TP / TN / FP / FN & 50 / 49 / 0 / 1 \\
\bottomrule
\end{tabular}
\caption{Validation of the LLM judge against human-majority labels on 100 open-ended QA examples.}
\label{tab:judge_validation}
\end{table}

\section{Case Study} \label{app:case}
\begin{figure*}[t]
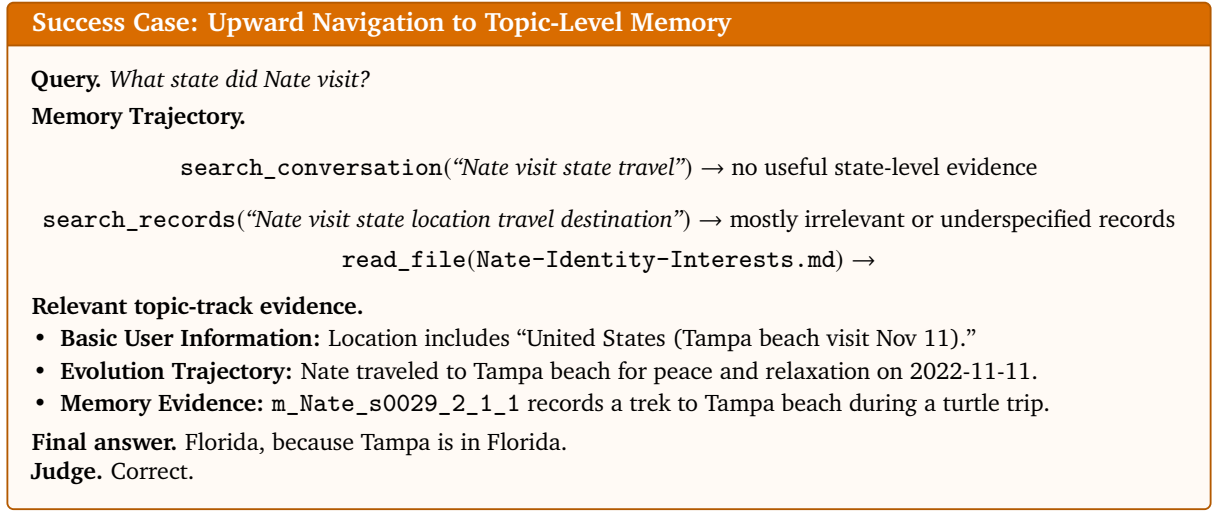

\centering
\begin{tcolorbox}[
    enhanced,
    colback=orange!4,
    colframe=orange!70!black,
    colbacktitle=orange!85!black,
    coltitle=white,
    title=Success Case: Upward Navigation to Topic-Level Memory,
    fonttitle=\bfseries,
    boxrule=0.7pt,
    arc=2pt,
    width=0.96\textwidth,
    left=6pt,
    right=6pt,
    top=6pt,
    bottom=6pt
]
\small

\textbf{Query.}
\textit{What state did Nate visit?}

\vspace{0.3em}
\textbf{Memory Trajectory.}
\[
\texttt{search\_conversation}(\textit{``Nate visit state travel''})
\rightarrow
\text{no useful state-level evidence}
\]
\[
\texttt{search\_records}(\textit{``Nate visit state location travel destination''})
\rightarrow
\text{mostly irrelevant or underspecified records}
\]
\[
\texttt{read\_file}(\texttt{Nate-Identity-Interests.md})
\rightarrow 
\]

\vspace{0.3em}
\textbf{Relevant topic-track evidence.}
\begin{itemize}[leftmargin=1.2em, itemsep=0.15em, topsep=0.1em]
    \item \textbf{Basic User Information:} Location includes ``United States (Tampa beach visit Nov 11).''
    \item \textbf{Evolution Trajectory:} Nate traveled to Tampa beach for peace and relaxation on 2022-11-11.
    \item \textbf{Memory Evidence:} \texttt{m\_Nate\_s0029\_2\_1\_1} records a trek to Tampa beach during a turtle trip.
\end{itemize}

\vspace{0.3em}
\textbf{Final answer.}
Florida, because Tampa is in Florida.

\textbf{Judge.} Correct.
\end{tcolorbox}
\caption{Success case.}
\label{fig:success_case}
\end{figure*}

Figure~\ref{fig:success_case} shows a successful case where upward navigation helps recover missing context. Low-level search over raw messages and records returns noisy or underspecified evidence, while the topic track summarizes a Tampa beach visit and points the agent to the correct answer. This case illustrates that higher-level memory can help when direct retrieval fails to surface the needed evidence.
Figure~\ref{fig:failure_case} shows a complementary failure mode. The agent retrieves relevant records, but the evidence is not strong enough to support the exact premise of the selected option. In particular, it over-generalizes an event-level memory about a successful salsa fundraiser into a stable preference for weekend salsa dance socials. Future navigation policies should better calibrate evidence strength, especially when converting event evidence into user preferences.

\begin{figure*}[t]
\centering
\begin{tcolorbox}[
    enhanced,
    colback=red!3,
    colframe=red!65!black,
    colbacktitle=red!75!black,
    coltitle=white,
    title=Failure Case: Over-generalizing Event Evidence into a Stable Preference,
    fonttitle=\bfseries,
    boxrule=0.7pt,
    arc=2pt,
    width=0.96\textwidth,
    left=6pt,
    right=6pt,
    top=6pt,
    bottom=6pt
]
\small

\textbf{Query.}
\textit{Do you have suggestions for peaceful weekend activities that can help me recharge and spark inspiration?}

\vspace{0.3em}
\textbf{Answer options.}
A. Since you already love spending weekends at lively salsa dance socials, you might enjoy adding a recharge element by strolling through the Pilsen murals in the afternoon, or sipping café de olla at a local spot while jotting down memories from your favorite dance nights.

B. Since you already love spending weekends on your favorite hobbies, you might enjoy adding a peaceful twist by taking morning walks in nature, or relaxing at a local café with a notebook, letting your thoughts wander into new ideas.

C. Since you already love spending weekends at silent creative writing retreats, you might enjoy extending that peace with sunrise walks along the lakefront, or taking a quiet afternoon to photograph hidden corners of the city. Pair that with journaling in a cozy coffee shop, letting the rhythms of the neighborhood inspire fresh story ideas.

D. Since you already love spending weekends experimenting with new vegan baking recipes, you might enjoy extending that creativity by visiting local farmers markets and chatting with growers about seasonal ingredients, or taking a peaceful afternoon to sketch in a sunny park while your latest bread rises at home."

\textbf{Memory trajectory.}
\[
\begin{aligned}
&\texttt{search\_records}(\textit{weekend activities, hobbies, salsa})\rightarrow\texttt{search\_records}(\textit{Gabriela, preferences, relaxation}) \\
\rightarrow\;&\texttt{get\_records}([\textit{\texttt{m\_persona\_893\_s0001\_3\_2\_2}}])\rightarrow\;\texttt{get\_records}(\textit{energy from packed events record}) \\
\rightarrow\;&\text{Answer A}
\end{aligned}
\]

\textbf{Retrieved evidence.}
\begin{itemize}[leftmargin=1.2em, itemsep=0.15em, topsep=0.1em]
    \item A record states that the user considered changing a fundraiser entertainment lineup to salsa because the previous year's salsa set was successful.
    \item Another record states that the user feels energized by packed big events and community connections, but also needs alone time afterward to recover.
    \item Other retrieved records include forgetting constraints or negative preferences, such as no longer liking certain solitary or outdoor activities.
\end{itemize}

\vspace{0.3em}
\textbf{Failure mode.}
The agent treats a successful salsa set at a fundraiser as evidence that the user personally loves weekend salsa dance socials. This over-generalizes an event-level memory into a stable preference and makes option A appear more strongly supported than the evidence warrants.

\vspace{0.3em}
\textbf{Takeaway.}
Relevant memories may still be too weak to support the exact premise of an answer option. Future navigation policies should better distinguish event evidence from stable preference evidence and avoid upgrading weak associations into personal traits.

\end{tcolorbox}
\caption{Failure case where relevant retrieved memories lead to over-generalization from an event-level memory to a stable user preference.}
\label{fig:failure_case}
\end{figure*}

\section{Prompt} \label{app:prompt}
\subsection{Judge Prompt}
This prompt is used for LoCoMo and LongMemEval judged accuracy during evaluation and RL reward computation.
\begin{tcolorbox}[
colback=orange!5,
colframe=orange!60,
title=Open-QA LLM Judge Prompt,
fonttitle=\bfseries,
boxrule=0.5pt,
arc=2pt,
left=4pt,
right=4pt,
top=4pt,
bottom=4pt
]
\small
\textbf{System:}

You are a strict but fair evaluator for open-domain QA.
Decide whether the predicted answer correctly answers the question given the ground truth answer.
Accept paraphrases, aliases, equivalent dates, and equivalent numbers.
Ignore extra explanation unless it contradicts the ground truth.
Output only 1 if the prediction is correct; output only 0 otherwise.

\textbf{User:}

Question:
\{question\}

Ground truth answer:
\{ground\_truth\_answer\}

Predicted answer:
\{predicted\_answer\}

Return only 1 or 0.
\end{tcolorbox}

\subsection{Memory Pyramid Prompts}
Figure~\ref{fig:memory_extract_prompt}--\ref{fig:profile_prompt} show the abbreviated prompt templates used to construct the memory pyramid, covering memory record extraction, record reconciliation, topic-track writing, and user-profile updating.

\subsection{Inference Prompts}
Figure~\ref{fig:navigation_prompt} shows the abbreviated prompt templates used to instruct the model to actively navigate the memory space.
\begin{figure*}[t]
\centering
\begin{tcolorbox}[
colback=black!5,
colframe=black!60,
title=Memory Record Extraction Prompt,
fonttitle=\bfseries,
boxrule=0.5pt,
arc=2pt,
width=0.96\textwidth,
left=5pt,
right=5pt,
top=5pt,
bottom=5pt
]
\small
You are a scene segmentation and memory extraction expert.

Given the previous scene, background messages, and newly appended messages, perform two tasks:

\textbf{Scene segmentation.}
Identify the current conversation scene. Inherit the previous scene if there is no clear topic transition; otherwise create a new concise scene name.

\textbf{Memory extraction.}
Extract memory records only from the newly appended messages. Each memory must be self-contained, grounded in source messages, and belong to one of four types:
\textit{fact}, \textit{event}, \textit{instruction}, or \textit{preference}.

Do not extract trivial small talk, temporary one-off requests without reusable information, repeated content, or unsupported subjective feelings. Preserve answerable details such as names, dates, numbers, prices, locations, status changes, and completed actions.

\textbf{Output.}
Return a JSON array. Each item contains a scene name, message ids, and extracted memories:
\begin{verbatim}
[
  {
    "scene_name": "...",
    "message_ids": ["..."],
    "memories": [
      {
        "content": "...",
        "type": "fact|event|instruction|preference",
        "priority": 80,
        "source_message_ids": ["..."],
        "metadata": {}
      }
    ]
  }
]
\end{verbatim}

Output only valid JSON and no extra text.
\end{tcolorbox}
\caption{Abbreviated prompt template for memory record extraction.}
\label{fig:memory_extract_prompt}
\end{figure*}

\begin{figure*}[t]
\centering
\begin{tcolorbox}[
colback=black!5,
colframe=black!60,
title=Memory Record Reconciliation Prompt,
fonttitle=\bfseries,
boxrule=0.5pt,
arc=2pt,
width=0.96\textwidth,
left=5pt,
right=5pt,
top=5pt,
bottom=5pt
]
\small
You are a memory conflict detector.

Given a set of newly extracted memory records and a retrieved candidate pool of existing records, decide how each new record should be handled.

\textbf{Actions.}
For each new record, choose one of:
\begin{itemize}[leftmargin=*]
    \item \texttt{store}: add the new record as new information.
    \item \texttt{skip}: discard the new record because it is redundant or less informative.
    \item \texttt{update}: replace older records when the new record is newer, more specific, or corrects outdated information.
    \item \texttt{merge}: combine complementary records that describe the same fact, state, event, instruction, or evolving process.
\end{itemize}

\textbf{Decision criteria.}
Compare records by subject, theme, time, and scene context. Cross-type merging is allowed when records describe the same underlying information. Preserve exact values such as names, dates, prices, quantities, item lists, locations, and status changes.

\textbf{Provenance.}
For \texttt{update} and \texttt{merge}, the final record must preserve the union of source message identifiers, timestamps, and metadata from the new and replaced records.

\textbf{Output.}
Return a JSON array, one decision per new record:
\begin{verbatim}
[
  {
    "record_id": "...",
    "action": "store|skip|update|merge",
    "target_ids": ["..."],
    "merged_content": "...",
    "merged_type": "fact|event|instruction|preference",
    "merged_priority": 85,
    "merged_timestamps": ["..."],
    "merged_source_message_ids": ["..."],
    "merged_metadata": {}
  }
]
\end{verbatim}
Output only valid JSON and no extra text.
\end{tcolorbox}
\caption{Abbreviated prompt template for memory record reconciliation.}
\label{fig:dedup_prompt}
\end{figure*}

\begin{figure*}[t]

\centering

\begin{tcolorbox}[
colback=black!5,
colframe=black!60,
title=Topic Track Writing Prompt,
fonttitle=\bfseries,
boxrule=0.5pt,
arc=2pt,
width=0.96\textwidth,
left=5pt,
right=5pt,
top=5pt,
bottom=5pt
]

\small

You are a memory consolidation writer.

Given newly committed memory records, the existing topic-track index, and the current timestamp, consolidate the records into coherent topic tracks.

\textbf{Goal.}

Build medium-range user contexts rather than a list of isolated records. Each topic track should summarize a recurring user topic, such as an ongoing project, preference domain, relationship, routine, or long-term goal.

\textbf{Actions.}

Prefer updating an existing topic track when possible. Create a new track only when the new records cannot be integrated into any existing topic. Merge related tracks when they describe the same topic or narrative arc. The total number of topic tracks should not exceed 20.

\textbf{Writing requirements.}

Each topic track should be written as a Markdown document with metadata, a short summary, a narrative description, and an evidence section. The narrative should integrate new and existing information coherently rather than appending records mechanically.

\textbf{Provenance.}

Every integrated memory must be grounded with its supporting memory record identifier in the evidence section.

\textbf{Output behavior.}

Use file operations to read, update, create, or merge topic tracks. Finish the round after all necessary edits are completed.

\end{tcolorbox}
\caption{Abbreviated prompt template for topic-track writing.}
\label{fig:topic_track_prompt}
\end{figure*}

\begin{figure*}[t]
\centering
\begin{tcolorbox}[
colback=black!5,
colframe=black!60,
title=User Profile Update Prompt,
fonttitle=\bfseries,
boxrule=0.5pt,
arc=2pt,
width=0.96\textwidth,
left=5pt,
right=5pt,
top=5pt,
bottom=5pt
]
\small
You are a user-profile synthesis writer.

Given the existing user profile, newly changed topic tracks, update statistics, and the current timestamp, revise the global user profile.

\textbf{Evidence constraints.}
Use only information grounded in topic tracks. Do not infer user information from file paths, workspace structure, system metadata, or unsupported assumptions. If evidence is insufficient, omit the field rather than speculate. Keep the output language consistent with the source records.

\textbf{Synthesis goals.}
Build a concise user-level narrative by connecting information across topic tracks. Focus on stable information rather than episodic details. Scan for:
\begin{itemize}[leftmargin=*]
    \item \textbf{Base anchors}: confirmed facts, demographic traits, and current state.
    \item \textbf{Interest graph}: topics, activities, or domains that receive sustained attention.
    \item \textbf{Interaction protocol}: communication style, workflow preferences, and response constraints.
    \item \textbf{High-level patterns}: recurring decision logic, long-term tendencies, and supported evolution.
\end{itemize}

\textbf{Output.}
Return a Markdown user profile within 2000 characters. The profile may contain sections for basic information, long-term preferences, current context, interaction protocol, and high-level evolution. Revise the existing profile rather than rewriting from scratch, preserving useful stable information and updating outdated content when new evidence supports the change.
\end{tcolorbox}
\caption{Abbreviated prompt template for user-profile updates.}
\label{fig:profile_prompt}
\end{figure*}

\begin{figure*}[t]
\centering
\begin{tcolorbox}[
colback=blue!4,
colframe=blue!50,
title=Inference Prompt,
fonttitle=\bfseries,
boxrule=0.5pt,
arc=2pt,
width=0.96\textwidth,
left=5pt,
right=5pt,
top=5pt,
bottom=5pt
]
\small
You are a personal assistant. Your goal is to answer the user's question correctly and grounded in evidence.

\textbf{Workflow.}
First, determine whether the current information is sufficient. If the question concerns the user and the evidence is insufficient, use the structured memory bank through tools. Stop using tools once enough evidence has been gathered.

\textbf{Tool budget.}
Use no more than four tool-calling turns per question. Do not continue searching for marginal extra evidence.

\textbf{Memory bank.}
The memory bank contains four layers:
\begin{itemize}[leftmargin=*]
    \item \textbf{Raw conversations}: original dialogue evidence; use \texttt{search\_conversation} or \texttt{get\_conversation}.
    \item \textbf{Memory records}: compact structured memories; use \texttt{search\_records} or \texttt{get\_records}.
    \item \textbf{Topic tracks}: topic-level evolving summaries; use \texttt{read\_file}.
    \item \textbf{User profile}: global user-level summary; use \texttt{read\_file}.
\end{itemize}

\textbf{Grounding.}
Do not guess user information. If retrieved memories conflict, strict instruction records take precedence. Use raw conversations when exact evidence is needed.

\textbf{Final answer.}
Only when evidence is sufficient, return the answer in the required JSON format:
\begin{verbatim}
{"answer": "...", "reason": "..."}
\end{verbatim}
\end{tcolorbox}
\caption{Abbreviated prompt template for inference-time memory navigation.}
\label{fig:navigation_prompt}
\end{figure*}

\subsection{Artifact Licenses}
\label{app:artifact_licenses}

Table~\ref{tab:artifact_licenses} summarizes the licenses of the external datasets, benchmarks, and model checkpoints used in this work. We use all artifacts for research and evaluation purposes, and follow the corresponding license terms and citation requirements.

\begin{table*}[t]
\centering
\small
\renewcommand{\arraystretch}{1.08}
\begin{tabular}{p{0.17\textwidth}p{0.18\textwidth}p{0.18\textwidth}p{0.38\textwidth}}
\toprule
\textbf{Artifact} & \textbf{Type} & \textbf{License} & \textbf{Use in this work} \\
\midrule
LoCoMo & Long-term conversational memory benchmark & CC BY-NC 4.0 & Used for long-term user-memory evaluation and part of memory-tool RL training. The license is non-commercial, so we use it for research evaluation only. \\
\midrule
PersonaMem-v2 & Personalized user-memory benchmark & MIT & Used for implicit preference and persona-memory evaluation, and part of memory-tool RL training. \\
\midrule
LongMemEval & Long-term memory benchmark & MIT & Used as a held-out out-of-distribution memory benchmark. \\
\midrule
V*Bench & Visual tool-use benchmark & MIT project license & Used for non-memory visual tool-use evaluation. The official repository is MIT-licensed; image assets should be used following their original source terms when redistributed. \\
\midrule
GPQA-Diamond & Scientific reasoning benchmark & MIT & Used as a non-memory reasoning benchmark. \\
\midrule
BFCL-v3 & Function-calling benchmark & Apache-2.0 & Used as a non-memory text tool-use benchmark. \\
\midrule
Qwen3.5-9B / 122B / 397B & Base language models & Apache-2.0 & Used as base LLMs for \method and scaling variants. \\
\bottomrule
\end{tabular}
\caption{Licenses of external artifacts used in this work.}
\label{tab:artifact_licenses}
\end{table*}

\end{document}